\pdfoutput=1

\documentclass[11pt]{article}

\usepackage[]{EMNLP2022}

\usepackage{graphicx}
\usepackage{tabularx}
\usepackage{soul}
\usepackage{xstring}
\usepackage{times}
\usepackage{latexsym}
\usepackage{hyperref}
\usepackage{amsfonts}
\usepackage{amsmath}
\usepackage{amssymb}
\usepackage{mathtools}
\usepackage{bm}
\usepackage{lscape}
\usepackage{algorithm,algpseudocode}
\usepackage{xcolor}
\usepackage[normalem]{ulem}
\usepackage{caption}
\usepackage{setspace}
\usepackage{booktabs}
\usepackage{enumitem}
\usepackage{multirow}
\usepackage{footnote}
\makesavenoteenv{tabular}
\makesavenoteenv{table}

\usepackage{inconsolata}

\usepackage{CJKutf8}
\usepackage[utf8]{inputenc}
\usepackage[T1]{fontenc}

\usepackage[T1]{fontenc}

\usepackage[utf8]{inputenc}

\usepackage{microtype}

\title{Domain Adaptation of Machine Translation with Crowdworkers}

\author{Makoto Morishita$^{1}$, Jun Suzuki$^{2}$, Masaaki Nagata$^{1}$ \\
  NTT Communication Science Laboratories, NTT Corporation$^{1}$ \\
  Tohoku University$^{2}$ \\
  \texttt{\{makoto.morishita.gr, masaaki.nagata.et\}@hco.ntt.co.jp} \\
  \texttt{jun.suzuki@tohoku.ac.jp}
  }

\begin{document}
\maketitle
\begin{abstract}
Although a machine translation model trained with a large in-domain parallel corpus achieves remarkable results, it still works poorly when no in-domain data are available.
This situation restricts the applicability of machine translation when the target domain's data are limited. However, there is great demand for high-quality domain-specific machine translation models for many domains.
We propose a framework that efficiently and effectively collects parallel sentences in a target domain from the web with the help of crowdworkers.
With the collected parallel data, we can quickly adapt a machine translation model to the target domain.
Our experiments show that the proposed method can collect target-domain parallel data over a few days at a reasonable cost.
We tested it with five domains, and the domain-adapted model improved the BLEU scores to +19.7 by an average of +7.8 points compared to a general-purpose translation model.
\end{abstract}

\section{Introduction}

\begin{figure}[t]
\centering
\includegraphics[width=0.9\linewidth]{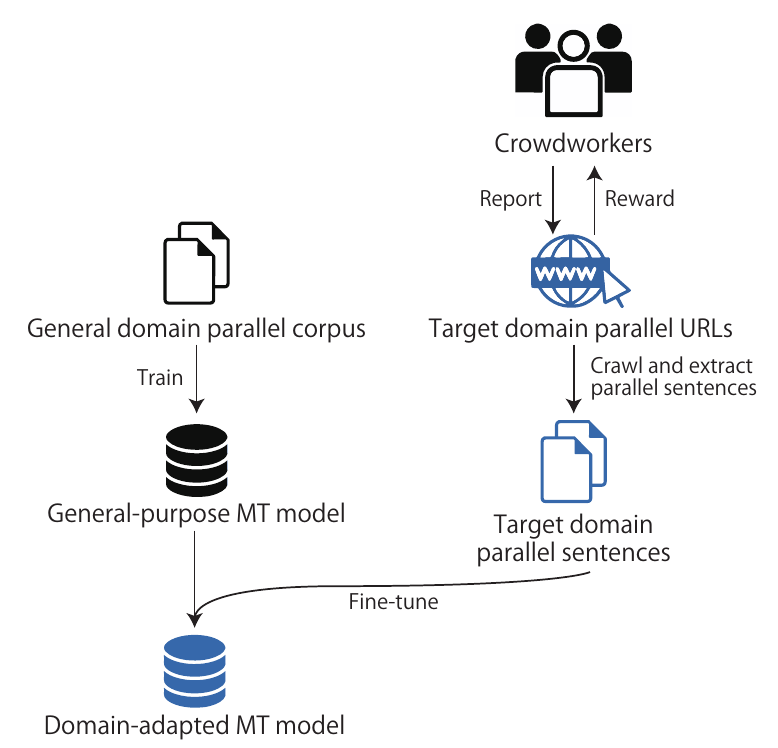}
\caption{Overview of proposed domain-adaptation method with crowdworkers who collected URLs that included parallel sentences of target domain. We then fine-tuned a general-purpose model with the collected target domain parallel sentences. See Section~\ref{sec:collecting_with_crowdworkers} for details.}\label{fig:overview}
\end{figure}

Although recent Neural Machine Translation (NMT) methods have achieved remarkable performance, their translation quality drastically drops when the input domain is not covered by training data~\cite{muller20amta}.
One typical approach for translating such inputs is adapting the machine translation model to a domain with a small portion of in-domain parallel sentences~\cite{chu18coling}.
Such sentences are normally extracted from a large existing parallel corpus~\cite{wang17acl,vanderwees17emnlp} or created synthetically from a monolingual corpus~\cite{chinearios17wmt}.
However, the existing parallel/monolingual data may not include enough sentences relevant to the target domain.

There is a real-world need for a method that can adapt a machine translation model to any domain. For example, users  reading or writing in such specific fields as scientific, medical or patent domains, may experience satisfaction if they have access to a domain-adapted machine translation model.
Unfortunately, the often limited availability of in-domain parallel data complicates this task.
For example, it is difficult to adapt a model to the COVID-19 domain because this issue is too new, and the current available data do not sufficiently cover it.

To alleviate the issue, we propose a method that rapidly adapts a machine translation model to many domains at reasonable costs and time periods with crowdworkers.
Fig.~\ref{fig:overview} shows an overview of our framework.
We hypothesize that a small number of in-domain parallel sentences of the target domain are available on the web, and we ask crowdworkers to report these web URLs as a web mining task.
Our task does not require translation skills,
unlike some previous research~\cite{zaidan11crowdsourcing,behnke18lrec,kalimuthu19incremental} that attempted manual translations of in-domain monolingual sentences by crowdworkers.
Thus, workers who are not professional translators can participate.

Furthermore, to collect effective parallel sentences, we also vary the crowdworkers' rewards based on the quality of their reported URLs.
After collecting parallel sentences by our method, we adapted the machine translation model with the collected, target-domain parallel sentences.
Our method has the advantage of being applicable to many domains, in contrast to previous works that use existing parallel/monolingual data.

We experimentally show that our method quickly collects in-domain parallel sentences and improves the translation performance of the target domains in a few days and at a reasonable cost.

Our contributions can be summarized as follows:
\begin{itemize}
    \item We proposed a new domain-adaptation method that quickly collects in-domain parallel sentences from the web with crowdworkers.
    \item We empirically showed that crowdworkers are motivated by variable rewards to find more valuable web sites and achieved better performance than under the fixed reward system.
\end{itemize}

\section{Related Work}

\subsection{Domain Adaptation}
Domain adaptation is a method that improves the performance of a machine translation model for a specific domain.
The most common method for neural machine translation models is to fine-tune the model with target-domain parallel sentences~\cite{chu18coling}.
\newcite{kiyono20wmt}, who ranked first in the WMT 2020 news shared task~\cite{barrault20wmt}, fine-tuned a model with a news domain parallel corpus and improved the BLEU scores by +2.2 points.
Since the availability of a target-domain parallel corpus is limited, we typically select similar domain sentences from a large parallel corpus~\cite{moore-lewis10acl,axelrod11emnlp}.
However, its applicability remains limited because some domains are not covered by existing parallel corpora.

We take a different approach that freshly collects target-domain parallel sentences from the web.
Since we do not rely on an existing corpus, our method can be applied to many domains.

\subsection{Collecting Parallel Sentences from the Web}

Recently, some works successfully built a large-scale parallel corpus by collecting parallel sentences from the web.
The BUCC workshop organized shared-tasks of extracting parallel sentences from the web~\cite{sharoff15bucc,zweigenbaum17bucc}.
The ParaCrawl project successfully created a large-scale parallel corpus between English and other European languages by extensively crawling the web~\cite{banon20paracrawl}.
Typical bitext-mining projects, including ParaCrawl, took the following steps to identify parallel sentences from the web~\cite{resnik03webasparallelcorpus}:
(1) find multilingual websites, which may contain parallel sentences, from the web~\cite{papavassiliou18lrec,banon20paracrawl};
(2) find parallel documents from websites~\cite{thompson20emnlp,elkishky20aacl};
(3) extract parallel sentences from parallel web URLs~\cite{thompson19vecalign,chousa20coling}.
Our work focuses on the first step: finding bilingual target-domain web URLs.
\newcite{banon20paracrawl} analyzed all of the CommonCrawl data to find crawl candidate websites that contain a certain amount of both source and target language texts.
Their method efficiently collected parallel sentences from the web. However, since CommonCrawl only covers a small portion of the web, it may overlook websites that contain valuable resources.
Thus, the current web-based corpora~\cite{banon20paracrawl,morishita20lrec} may not cover all the domains we want to adapt.
It is also difficult to focus on a specific topic. In contrast, our work does not rely on CommonCrawl but on crowdworkers who can search the whole web and focus on specific domains.

\subsection{Creating Parallel Corpus with Crowdworkers}
Some researchers have used crowdsourcing platforms to create new language resources~\cite{roit20acl,jiang18naacl}.
Some work created a parallel corpus for domain-adaptation by asking crowdworkers to translate in-domain monolingual sentences \cite{zaidan11crowdsourcing,behnke18lrec,kalimuthu19incremental}.
Although this approach is straightforward, it does suffer from several drawbacks.
For example, it is often difficult to find a sufficient amount of crowdworkers since translation tasks often require an understanding of both the languages that are actually being used.
Note that although we also use a crowdsourcing platform, our approach entirely differs from the approach introduced in this section, such as asking crowdworkers to do translation tasks.

\section{Collecting Parallel URLs with Crowdworkers}
\label{sec:collecting_with_crowdworkers}
Fig.~\ref{fig:overview} shows an overview of our collecting protocol.
Our method asks workers to find URLs that are related to the target domain and written in parallel.
We then extract the parallel sentences from these URLs and fine-tune the general-purpose machine translation model with the collected data.

This section is organized as follows:
In Section~\ref{sec:advantages}, we explain why we focus on collecting parallel URLs and describe their advantages.
We overview the details of our crowdsourcing task definition in Section~\ref{sec:task_definition}.
In Section~\ref{sec:sentence_extraction}, we describe how we extract parallel sentences from the reported URLs.
We describe the details of our reward setting in Section~\ref{sec:reward}.

\subsection{Advantages}
\label{sec:advantages}

Previous works, which adapted a machine translation model to a specific domain, created resources by asking crowdworkers to translate text~\cite{lewis11wmt,anastasopoulos20tico19,zaidan11crowdsourcing,behnke18lrec,kalimuthu19incremental}.
In contrast, our method asks workers to find web URLs (instead of translating sentences) that have parallel sentences in the target domain.
This method has two advantages.
The first concerns task difficulty.
To achieve rapid domain adaptation, the task must be easy enough that many crowdworkers can participate.
Thus, we do not assume that the workers fluently understand both the source and target languages.
Finding potential web URLs that have parallel sentences is relatively easy and can be done by any crowdworker.

The other advantage involves task efficiency.
We asked workers to collect the URLs of parallel web pages instead of parallel sentences
because recent previous works successfully extracted parallel sentences from parallel URLs~\cite{banon20paracrawl}.
Efficiency is important for our method, since we focus on speed to create a domain-specific model.

\subsection{Crowdsourcing Task Definition}
\label{sec:task_definition}
We focus on collecting the parallel sentences of languages $e$ and $f$.
We created a web application to accept reports from the crowdworkers and extracted parallel sentences from the reported web URLs.
We prepared a development set (a small portion of the parallel sentences) of the target domain and distribute it to the workers as examples of the type of sentences we want them to collect.
The crowdworkers are asked to find pairs of web URLs that contain parallel sentences of the target domain.
We call this URL pair a parallel URL.
Note that we collect the URLs of pages written in parallel; this means that workers act as parallel document aligners.
We do not accept parallel URLs that have already been reported by others.

\subsection{Parallel Sentence Extraction}
\label{sec:sentence_extraction}

After obtaining parallel URLs from workers, we extract parallel sentences from the reported URLs.
First, we downloaded the reported web URLs and extracted the texts\footnote{Since we expect the workers to act as document aligners, we focus on the reported URLs and do not crawl the links in the reported URLs.}
and removed the sentences that are not in the $e$ or $f$ language based on {\tt CLD2}\footnote{\url{https://github.com/CLD2Owners/cld2}}.
Then we used {\tt vecalign}~\cite{thompson19vecalign} to extract the parallel sentences, a step that
aligns them based on the multi-lingual sentence embeddings LASER~\cite{artetxe19laser}.
We discard noisy sentence pairs based on sentence alignment scores\footnote{Since {\tt vecalign} outputs a scoring cost where a lower score means better alignment, our implementation removes a sentence pair if its cost exceeds 0.7.} and do not use them for model training.

\subsection{Reward Settings}
\label{sec:reward}
To bolster the crowdworkers' motivation, reward setting is one of the most important issues~\cite{posch19acm}.
In this paper, we tested two types of rewards: fixed or variable.
In the following, we describe both reward settings.

\subsubsection{Fixed Reward}
Fixed reward pays a set amount for each reported URL if we can extract at least one parallel sentence from it.
This fixed reward setting is one very typical setting for crowdsourcing.

\subsubsection{Variable Reward}
\label{sec:variable_reward}
The key motivation of crowdworkers is probably to earn money~\cite{antin12chi}, and thus they try to maximize their earnings~\cite{horton10ec}.
Since the fixed reward setting only considers the number of reported URLs, workers may report noisy URLs whose texts are not parallel or not in the target domain in an effort to maximize their number of reports.

To alleviate this concern, we tested another reward setting: varying rewards based on the quality of their reported parallel URLs.
We hypothesize that the workers will improve their work performance when we pay more for good work and less for poor work.

We defined parallel URLs as those satisfying the following criteria that help improve the translation performance in the target domain: (1) they contain a large number of parallel sentences, (2) the parallel sentences are correctly translated, and (3) the parallel sentences are in the target domain.
To reflect these criteria in the reward, we set variable reward $r$:
\begin{equation}
    \label{eq:reward}
    r = \min(r_{\rm max}, r_{\rm min} + \sum_{(x_{i}, y_{i}) \in \mathbb{D}} S_{a}(x_{i}, y_{i}) + S_{d}(x_{i})),
\end{equation}
where $\mathbb{D}$ is a set of parallel sentences extracted from the reported URLs, $x_{i}$ and $y_{i}$ are parallel sentences of languages $e$ and $f$, $r_{\rm min}$ and $r_{\rm max}$ are the minimum and maximum reward per report, and $S_{a}(\cdot)$ and $S_{d}(\cdot)$ are the sentence alignment and domain similarity scores, which are explained below.

\paragraph{Sentence Alignment Score}
Suppose $n$ parallel sentences $\mathbb{D} = \{(x_{1}, y_{1}), \dots, (x_{n}, y_{n})\}$ extracted from the reported URLs.
Sentence alignment score $S_{a}$ is calculated as follows:
\begin{equation}
    \label{eq:sentence_alignment}
    S_{a} = \sum_{(x_{i}, y_{i}) \in \mathbb{D}} \varsigma(-\mathrm{V}(x_{i}, y_{i})),
\end{equation}
where $\mathrm{V}(\cdot)$ is an alignment cost function of {\tt vecalign}, where lower is better, and $\varsigma(\cdot)$ is a sigmoid function that converts the score into the range 0 to 1.

\paragraph{Domain Similarity Score}
The domain similarity score is based on cross-entropy~\cite{moore-lewis10acl}:
\begin{equation}
    \label{eq:domain_similarity}
    S_{d} = \sum_{x_{i} \in \mathbb{D}} \varsigma(\mathit{H}_{I}(x_{i}) - \mathit{H}_{N}(x_{i})),
\end{equation}
where I and N are in-domain and non-domain-specific language models and $\mathit{H}(x_{i})$ is the per-word cross-entropy of sentence $x_{i}$.

Through our web application, workers can check the results (of their previous reports), which include the reward amounts, the scores, and the number of extracted parallel sentences.
These results are available a few minutes after we accept their reports so that they can improve their work and maximize their scores and their payments.

\section{Experiments}
\label{sec:large_exp}
We carried out experiments to confirm whether different reward settings influenced the workers' performance and translation accuracy.
Prior to them, we conducted a preliminary experiment to check the effect of our method in smaller settings.
Refer to Section~\ref{sec:small_exp} in the Appendix for this preliminary experiment.
In this section, we empirically confirm the effectiveness of our method by focusing on five domains.

\subsection{Experimental Settings}
\label{sec:large_exp_settings}
In this experiment, we tested English-Japanese translations on five domains: COVID-19, news, science, patents, and legal matters\footnote{We chose these domains because they require special domain knowledge and are difficult to translate by current models.}.
The details of the domains and the corpus statistics of the development/test sets are shown in Section~\ref{sec:appendix_datasets} in the Appendix.
We hired 97 crowdworkers through a crowdsourcing platform called Crowdworks\footnote{\url{https://crowdworks.jp/}}.
Each worker was randomly assigned to a single target domain.

We used both the fixed and variable reward setting for the science and patent domains, and only the variable reward setting for the other three domains, since we confirmed that the variable reward setting is effective in the following experiment (see Section~\ref{sec:large_reward}).
We set the fixed reward at 25 JPY ($\simeq 0.23$ USD), $r_{\rm min}$ to 10 JPY ($\simeq 0.09$ USD), and $r_{\rm max}$ to 100 yen ($\simeq 0.91$ USD) for the variable reward\footnote{We paid the workers in JPY since they mainly live in Japan. They are guaranteed at least the minimum wage.}.
Since our task is much easier than translating sentences, we pay our workers much less than such translators of sentences\footnote{Typically, it requires around 0.15 USD to translate an English word into Japanese.}.
Data collection continued for 13 days.
We trained in-domain language models with each development set to calculate the domain similarity scores.
We used {\tt KenLM} as an implementation of the $n$-gram language model~\cite{heafield11kenlm} to calculate the domain similarity scores.
We trained the in-domain language model with the development set of TICO-19 and the non-domain-specific model with JParaCrawl v2.0~\cite{morishita20lrec}.

\paragraph{Translation Model Settings}
As a neural machine translation model, we employed the Transformer model with its base settings~\cite{vaswani17transformer}.
To train the general-purpose baseline model, we used JParaCrawl v2.0~\cite{morishita20lrec}, which contains 10 million English-Japanese parallel sentences and
tokenized the training data into subwords with the {\tt sentencepiece}~\cite{kudo18sentencepiece} toolkit.
We set the vocabulary size to 32,000 for each language side and
removed sentences that exceeded 250 subwords to reduce the noisy sentence pairs.

We trained the baseline model with JParaCrawl until it converged and then fine-tuned it with the newly collected in-domain parallel sentences.
See Section~\ref{sec:appendix_hyperparameters} in the Appendix for the detailed hyperparameter settings.

We used {\tt SacreBLEU}~\cite{post18sacrebleu} to evaluate the translation performance and report the BLEU scores\footnote{We used NFKC to normalize both the Japanese translations and references since JParaCrawl is normalized by the same procedure.}~\cite{papineni02bleu}.

\subsection{Experimental Results}

\subsubsection{Fixed or Variable Reward Comparison}
\label{sec:large_reward}

\begin{figure}[t]
\centering
\includegraphics[width=\linewidth]{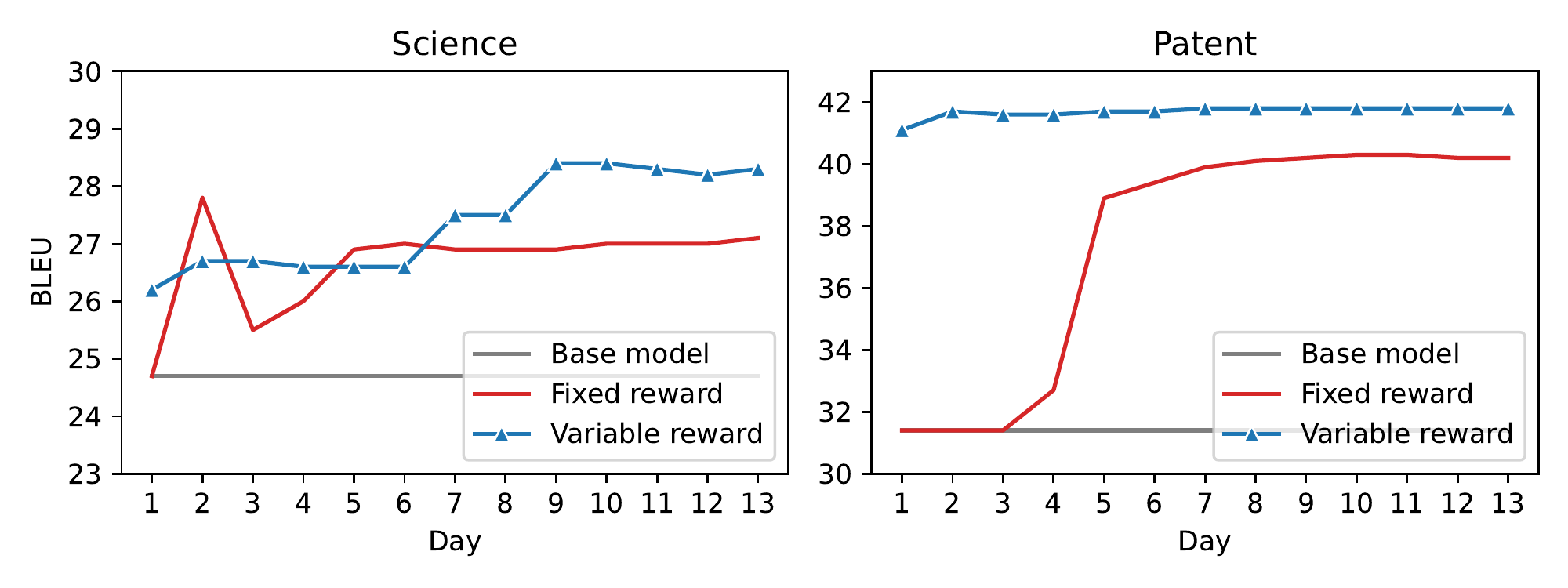}
\caption{Transition of test set BLEU scores on science and patent domains}\label{fig:large_reward_diff}
\end{figure}

First, we address whether the variable reward setting encouraged the workers to find more valuable data.
Fig.~\ref{fig:large_reward_diff} compares the BLEU scores between the fixed and variable reward settings in the science and patent domains.
The variable reward setting achieved higher BLEU scores than the fixed reward setting in both domains.
Combined with the preliminary experiments described in Section~\ref{sec:small_exp} in the Appendix, we conclude that the variable reward setting collects beneficial data.
Thus, the following sections mainly discuss the results of the variable reward setting.

\begin{table*}[t]
\centering
\scriptsize
\begin{tabular}{lrrrrrrrrrr}
\toprule
 & & & & \multicolumn{3}{c}{\bf Development BLEU} & \multicolumn{3}{c}{\bf Test BLEU} \\ \cmidrule(lr){5-7}\cmidrule(lr){8-10}
{\bf Domain} & {\bf \#URLs} & {\bf \#Sentences} & {\bf Cost (USD)} & {\bf Base model} & \multicolumn{2}{c}{\bf w/Crawled} & {\bf Base model} & \multicolumn{2}{c}{\bf w/Crawled} \\ \midrule
COVID-19 & 6,841 & 165,838 & 1,807.7 & 25.9 & {\bf 28.7} \hspace{-1.0em} & (+2.8) & 31.7 & {\bf 34.3} \hspace{-1.0em} & (+2.6)\\
News & 10,712 & 220,559 & 2,765.5 & 19.3 & {\bf 21.2} \hspace{-1.0em} & (+1.9) & 20.5 & {\bf 23.1} \hspace{-1.0em} & (+2.6)\\
Science & 10,948 & 390,303 & 3,217.8 & 25.0 & {\bf 27.9} \hspace{-1.0em} & (+2.9) & 24.7 & {\bf 28.3} \hspace{-1.0em} & (+3.6)\\
Patent & 4,135 & 307,104 & 1,431.3 & 27.4 & {\bf 36.6} \hspace{-1.0em} & (+9.2) & 31.4 & {\bf 41.8} \hspace{-1.0em} & (+10.4)\\
Legal & 5,438 & 302,747 & 2,088.0 &  22.9 & {\bf 42.0} \hspace{-1.0em} & (+19.1) & 22.8 & {\bf 42.5} \hspace{-1.0em} & (+19.7)\\
\bottomrule
\end{tabular}
\caption{Experimental results for five domains.
Model fine-tuned with newly crawled data significantly improved BLEU scores on all of them.
}\label{tab:large_exp_bleu}
\end{table*}

\subsubsection{Data Collection}
Table~\ref{tab:large_exp_bleu} shows the experimental results on the variable reward setting, including the number of URLs and collected parallel sentences.
Our framework collected a large number of parallel sentences for all five domains.
The lower half of Fig.~\ref{fig:large_exp_day_sent_bleu_graph} shows the transitions of the number of sentences collected with crowdsourcing on the COVID-19, news, and legal domains.
For the other domains, see Fig.~\ref{fig:large_exp_day_sent_bleu_graph_appendix} in the Appendix.
The number of collected sentences linearly increased as we continued crowdsourcing.

We carried out the task on the five domains and assigned roughly the same number of workers to each task, but we found that the number of reports differed.
This implies that the task’s difficulty might differ depending on the target domain.
For example, the science task might be easier than the others because several scientific journals translate abstracts (and make them available on the web) into other languages.

\subsubsection{Translation Performance}
\begin{figure*}[t]
\centering
\includegraphics[width=0.7\linewidth]{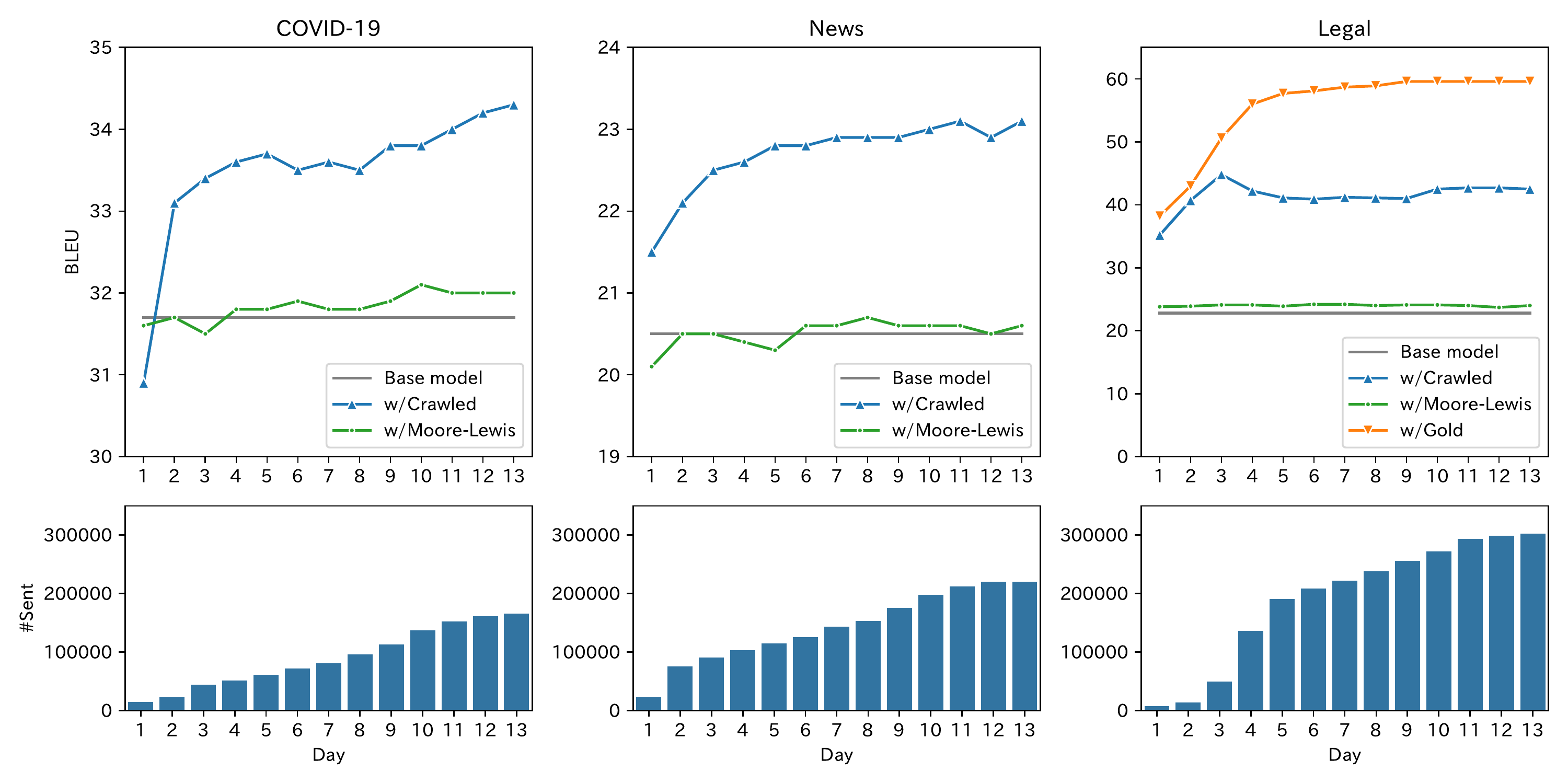}
\caption{
Transition of BLEU scores (top) and  sentences collected (bottom) as we continued data collection for the COVID-19, news, and legal domains.
Model named w/Moore-Lewis is fine-tuned with domain-relevant sentences extracted from existing general-purpose corpus, as described in Section~\ref{sec:moore-lewis}.
As an upper bound of fine-tuning, we show the scores of the w/Gold model, which was fine-tuned with existing target-domain parallel corpus, as described in Section~\ref{sec:gold} in the Appendix.
}\label{fig:large_exp_day_sent_bleu_graph}
\end{figure*}

Table~\ref{tab:large_exp_bleu} shows the BLEU scores of the baseline and the fine-tuned models with the collected in-domain parallel sentences.
The fine-tuned models achieved significantly better accuracy with an average of +7.8 points than the baseline model on all five domains.
In particular, our legal domain model improved by +19.7 points.
One likely reason is that the legal domain frequently uses words that do not appear in other domains, and the collected in-domain data improved these translations.

The top of Fig.~\ref{fig:large_exp_day_sent_bleu_graph} shows the transitions of the BLEU scores as we continued the data collection for the COVID-19, news, and legal domains (see the Base Model and w/Crawled lines).
Fig.~\ref{fig:large_exp_day_sent_bleu_graph_appendix} in the Appendix shows the results of the other domains.
All the domains show identical tendencies. Their performance surpassed the baseline on the first or second day of crowdsourcing and continued growing as we collected more data.
This supports our assumption that our method can achieve rapid domain adaptation for many domains.

\subsection{Comparison: Selecting In-domain Data from Existing Parallel Corpus}
\label{sec:moore-lewis}

\begin{table}[t]
\centering
\scriptsize
 \tabcolsep=3pt
\begin{tabular}{lrrrrr}
\toprule
 & \multicolumn{5}{c}{\bf Test BLEU}\\ \cmidrule(lr){2-6}
{\bf Domain} & {\bf Base} & \multicolumn{2}{c}{\bf w/Crawled} & \multicolumn{2}{c}{\bf w/ML} \\ \midrule
COVID-19 & 31.7 & 34.3 \hspace{-0.5em} & (+2.6) & 32.0 \hspace{-0.5em} & (+0.3)  \\
News & 20.5 & 23.1 \hspace{-0.5em} & (+2.6) & 20.6 \hspace{-0.5em} & (+0.1) \\
Science & 24.7 & 28.3 \hspace{-0.5em} & (+3.6) & 25.3 \hspace{-0.5em} & (+0.6) \\
Patent & 31.4 & 41.8 \hspace{-0.5em} & (+10.4) & 32.0 \hspace{-0.5em} & (+0.6) \\
Legal & 22.8 & 42.5 \hspace{-0.5em} & (+19.7) & 24.0 \hspace{-0.5em} & (+1.2) \\
\bottomrule
\end{tabular}
\caption{BLEU score comparisons with Moore-Lewis (w/ML)}\label{tab:large_exp_bleu_moore-lewis}
\end{table}

In this section, we compare our method with the existing domain adaption method
to answer the following question: Do we really need to collect new data with crowdworkers?

Currently, the most common domain-adaptation method is to find target domain sentences from existing parallel corpora~\cite{chu18coling}.
As with the existing method, we used the one proposed by~\newcite{moore-lewis10acl}\footnote{Some may be concerned that this method is outdated, but it is still considered a strong domain-adaptation method, since the recent first-ranked system among WMT submissions uses it for selecting relevant data~\cite{microsoft18wmt}.}.
We scored all the sentences in JParaCrawl and used those considered most relevant to the target domain.
We selected the same number of sentences as in our collected data.

Table~\ref{tab:large_exp_bleu_moore-lewis} shows the BLEU scores of each model, and the top of Fig.~\ref{fig:large_exp_day_sent_bleu_graph} shows the transition of the BLEU scores (see w/Moore-Lewis).
The Moore-Lewis method surpassed the baseline on all five domains, but by a narrow margin.
Although their method does not require additional cost, our method achieved significantly better performance with just a small additional cost.
Thus the answer to the above question is yes: our method outperformed the existing domain-adaptation method.

\section{Conclusion}
We introduced a new framework for domain adaptation in machine translation.
Our method asks crowdworkers to find parallel URLs related to the target domain.
Such a task does not require any professional skills and can be done cheaply by many people.
We then fine-tuned the machine translation model with parallel sentences in the target domain extracted from the reported URLs.
Through experiments, we empirically confirmed that our framework significantly improved the translation performance for a target domain within a few days of crowdsourcing and at a reasonable cost.
We also confirmed that our variable reward function, which is based on the quality of parallel sentences, changed the behavior of the workers who began to collect more effective parallel sentences, increasing the translation accuracy.

\section*{Limitations}
We assume that websites containing in-domain parallel sentences are available on the web, which might not be true for some difficult domains.
However, since we believe that parallel sentences in neighboring domains are available on the web, we expect our method to improve the translation accuracy on these domains.

We conducted English-Japanese experiments.
We expect our method to work on most major language pairs, including German-English and Chinese-Japanese, since there are many parallel websites on these language pairs.
However, we haven’t yet confirmed whether it does works on very minor language pairs, because finding parallel websites for them is difficult.

\section*{Ethics Statement}
In the experiments, our crawler strictly followed the ``robots.txt'' and crawled only from allowed websites.
During the experiments, we also ensured that the crowdworkers earned at least the minimum wage.

\section*{Acknowledgements}
We thank the three anonymous reviewers for their insightful comments.

\bibliography{myplain,main}
\bibliographystyle{acl_natbib}

\clearpage
\appendix

\section{Detailed Experimental Settings}

\subsection{Hyperparameters}
\label{sec:appendix_hyperparameters}

\begin{table}[t]
\centering
\scriptsize
\tabcolsep=3pt
\begin{tabular}{lp{43mm}}
\toprule
\multicolumn{2}{c}{\textbf{Base model}} \\ \midrule
Architecture           &   Transformer (base) \\
Optimizer              &   Adam ($\beta_{1}=0.9, \beta_{2}=0.98, \epsilon=1\times10^{-8}$)~\cite{kingma14adam}  \\
Learning rate schedule &   Inverse square root decay     \\
Warmup steps           &   4,000  \\
Max learning rate      &   0.001  \\
Dropout                &   0.3~\cite{srivastava14dropout} \\
Gradient clipping      &   1.0  \\
Label smoothing        &   $\epsilon_{ls}=0.1$~\cite{szegedy16cvpr}     \\
Mini-batch size        &   320,000 tokens  \\
Updates      &   24,000 updates\\
Averaging              &   Save checkpoint every 200 steps and average the last eight \\
Implementation         &   {\tt fairseq}~\cite{ott19fairseq} \\
Parameters   &   93.2 million \\ \midrule
\multicolumn{2}{c}{\textbf{Fine-tuning}} \\ \midrule
Learning rate          &   $1\times10^{-5}$ (Fixed)     \\
Mini-batch size        &   32,000 tokens  \\
Updates      &   8 epochs\footnote{One epoch means the model sees the entire corpus once. Thus the number of updates depends on the data size. We chose this setting because a fixed number of updates has a risk of over-fitting if the fine-tuning data are too small.}\\
Averaging              &   Save checkpoint every epoch and average the last eight \\
\bottomrule
\end{tabular}
\caption{List of hyperparameters}\label{tab:hyper-parameter}
\end{table}

Table~\ref{tab:hyper-parameter} shows the hyperparameter settings used to train a general-purpose machine translation model and fine-tune it with target domain sentences.
We did not conduct a hyperparameter search, and almost all the settings were borrowed from previous works ~\cite{morishita20lrec,kiyono20wmt}.

\subsection{Datasets}
\label{sec:appendix_datasets}

\begin{table}[t]
\centering
\scriptsize
\tabcolsep=3pt
\begin{tabular}{lrrrr}
\toprule
 & \multicolumn{2}{c}{\bf Development} & \multicolumn{2}{c}{\bf Test} \\ \cmidrule(lr){2-3}\cmidrule(lr){4-5}
{\bf Domain} & {\bf \#Sentences} & {\bf \#Tokens} & {\bf \#Sentences} & {\bf \#Tokens} \\ \midrule
COVID-19 & 971 & 21,085 & 2,100 & 49,490 \\
News & 1,998 & 45,318 & 1,000 & 22,141 \\
Science & 1,790 & 39,377 & 1,812 & 39,573 \\
Patent & 2,000 & 60,312 & 2,300 & 71,847 \\
Legal & 1,313 & 46,922 & 1,310 & 46,842 \\
\bottomrule
\end{tabular}
\caption{Number of sentences and English tokens in development and test sets}\label{tab:test_dev_corpus_stats}
\end{table}

We used TICO-19~\cite{anastasopoulos20tico19} as development and test sets to evaluate the translation performance of the COVID-19 domain.
Since the original TICO-19 does not include Japanese translations, professional translators translated the English sentences to create a Japanese reference. We used the development/test sets from the WMT20 news shared task~\cite{barrault20wmt} for the news domain and the NTCIR-10 patent translation task for the patent domain.
For the science domain, we used ASPEC~\cite{nakazawa16aspec}, which contains excerpts of scientific papers.
For the legal domain, we used the Japanese-English legal parallel corpus\footnote{\url{http://www.phontron.com/jaen-law/index.html}}.
Since it is not divided into development and test sets, we created them by randomly choosing sentences from the entire corpus.
The details of the development and test set corpus statistics are shown in Table~\ref{tab:test_dev_corpus_stats}.

\section{Preliminary Experiments}
\label{sec:small_exp}
We carried out a preliminary experiment to determine how the different reward settings influenced the workers' performance and translation accuracy.

\subsection{Experimental Settings}
\paragraph{Target Domain and Crowdsourcing Settings}

\begin{table*}[t]
\centering
\scriptsize
\begin{tabular}{lrrrrrrrrrr}
\toprule
 & & & & \multicolumn{3}{c}{\bf Development BLEU} & \multicolumn{3}{c}{\bf Test BLEU} \\ \cmidrule(lr){5-7}\cmidrule(lr){8-10}
{\bf Reward} & {\bf \#URLs} & {\bf \#Sentences} & {\bf Cost (USD)} & {\bf Base model} & \multicolumn{2}{c}{\bf w/Crawled} & {\bf Base model} & \multicolumn{2}{c}{\bf w/Crawled} \\ \midrule
Fixed & 504 & 5,220 & 322.8 & \multirow{2}{*}{25.9} & 26.3 \hspace{-1.0em} & (+0.4) & \multirow{2}{*}{31.7} & 31.9 \hspace{-1.0em} & (+0.2) \\ Variable & 503 & 6,722 & 284.3 & & {\bf 27.1} \hspace{-1.0em} & (+1.2) & & {\bf 33.2} \hspace{-1.0em} & (+1.5) \\ \bottomrule
\end{tabular}
\caption{Small-scale experiment's results (five days of crowdsourcing), including crowdsourcing results and BLEU scores of baseline and model fine-tuned with newly collected in-domain corpus.
}\label{tab:small_exp_bleu}
\end{table*}

In this experiment, we focused on English-Japanese translations in the COVID-19 domain.
We assigned ten crowdworkers to each reward setting and asked them to find websites that contained parallel sentences related to the COVID-19 domain.
The crowdsourcing continued for five days.

We set the fixed reward at 70 JPY ($\simeq 0.64$ USD) per report. For the variable reward setting, we paid $r$ JPY for each report, as shown by Eq.~\ref{eq:reward}.
We set $r_{\rm min}$ to 20 JPY ($\simeq 0.18$ USD) and $r_{\rm max}$ to 100 JPY($\simeq 0.91$ USD). Other model training settings, including the hyperparameters, are identical as in Section~\ref{sec:large_exp_settings}.

\subsection{Experimental Results}

\begin{figure}[t]
\centering
\includegraphics[width=0.7\linewidth]{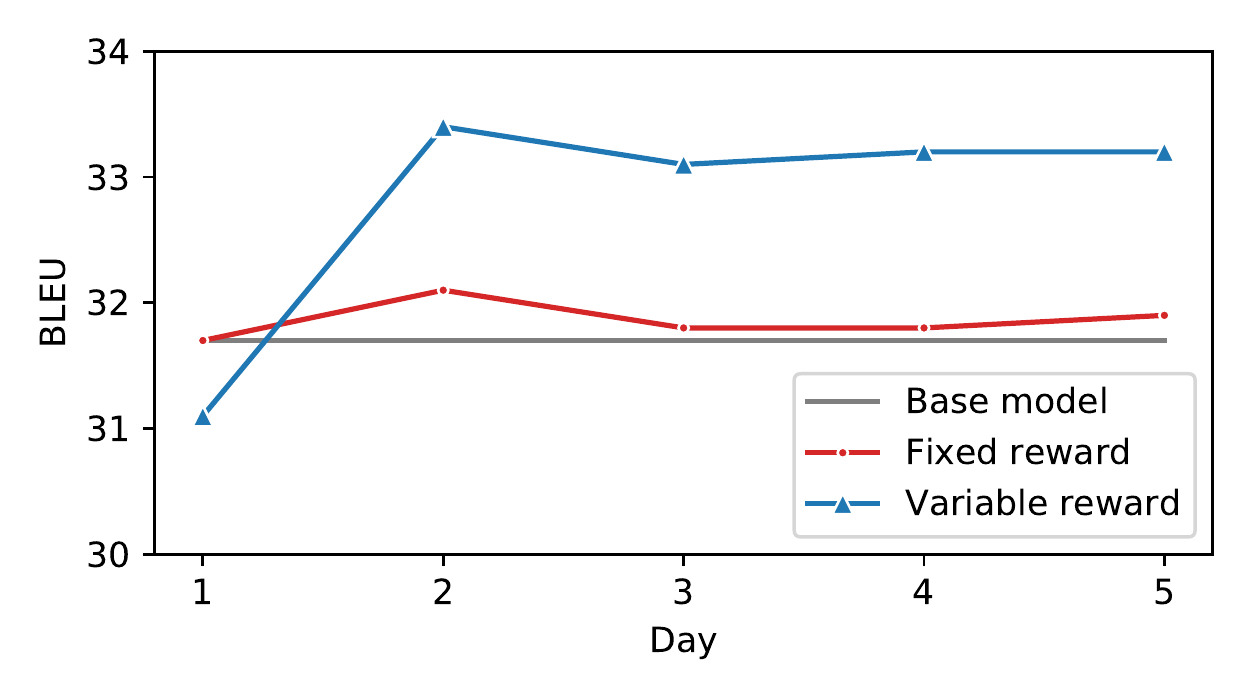}
\caption{Relationship between BLEU scores and crowdsourcing days for small-scale experiment}\label{fig:small_exp_day_bleu_graph}
\end{figure}

\subsubsection{Data Collection}
Table~\ref{tab:small_exp_bleu} shows the results of crowdsourcing, including the number of reports, extracted parallel sentences, and the payments to the workers.
We received almost the same number of reports in both reward settings.
However, there was a significant difference in the average number of sentences per report: 10.4 for the fixed rewards and 13.4 for the variable rewards.
One likely reason is that the workers tried to maximize their rewards.
We believe the number of in-domain parallel sentences is one crucial key for improving accuracy, and we reflected this idea in our reward function.
Thus it improved the workers' performance more than the fixed reward setting.
With the variable reward setting, we also reduced the cost and obtained even more parallel sentences
by reducing the payments to low-quality workers and increasing them to good workers.

\subsubsection{Translation Performance}
Table~\ref{tab:small_exp_bleu} shows the BLEU scores of the baseline model and the fine-tuned model with our crawled in-domain parallel data.
The model fine-tuned with variable reward data achieved better results than using fixed rewards.
We believe the quality of the collected data caused the difference in addition to the number of parallel sentences, as previously mentioned.
We compared the domain similarity scores described in Section~\ref{sec:variable_reward} to check whether the collected data are related to the target domain and found that the data collected with the variable reward setting achieved higher scores than with the fixed rewards.
This implies that the variable reward setting motivated the workers to find parallel web URLs related to the target domain, increasing the accuracy of the fine-tuned model.

Fig.~\ref{fig:small_exp_day_bleu_graph} shows how the BLEU scores changed as crowdsourcing continued, and Fig.~\ref{fig:small_exp_day_sent_graph} in the Appendix shows the number of sentences used for this experiment.
The fine-tuned model with the variable reward data outperformed the baseline model, even by the second day of crowdsourcing.
This result supports our claim that our method helps provide a domain-adapted model in a few days, which is critical in such urgent situations as COVID-19.

From this experiment, we found that a variable reward setting encouraged workers to find more valuable parallel URLs, improved their translation performance in the target domain over a few days, and reduced the cost more than the fixed reward setting.

\section{Additional Experimental Results}
\label{sec:appendix_exp_results}

\begin{figure}[t]
\centering
\includegraphics[width=\linewidth]{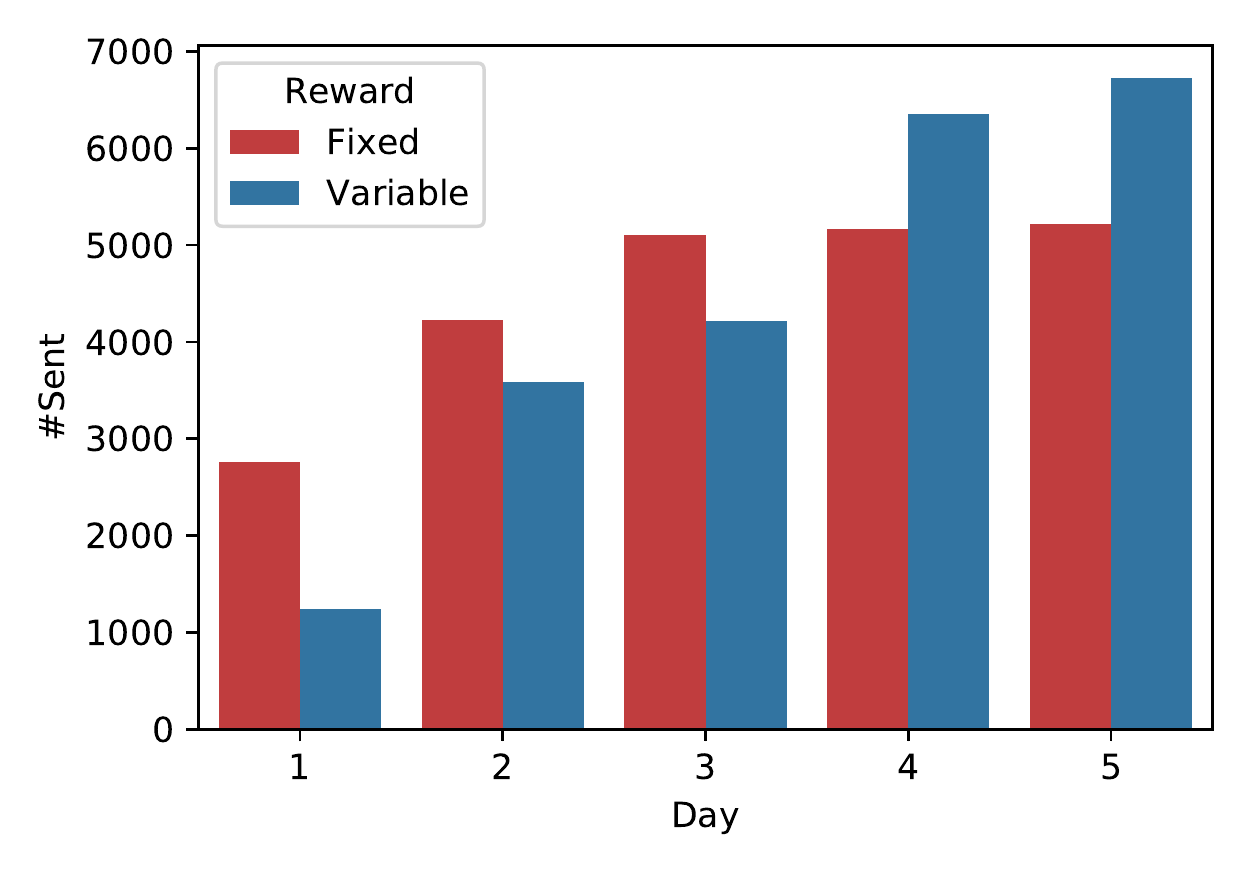}
\caption{Collected sentences used for fine-tuning in experiment of Fig.~\ref{fig:small_exp_day_bleu_graph}. See Section~\ref{sec:small_exp} for details.}\label{fig:small_exp_day_sent_graph}
\end{figure}

\begin{figure*}[t]
\centering
\includegraphics[width=0.8\linewidth]{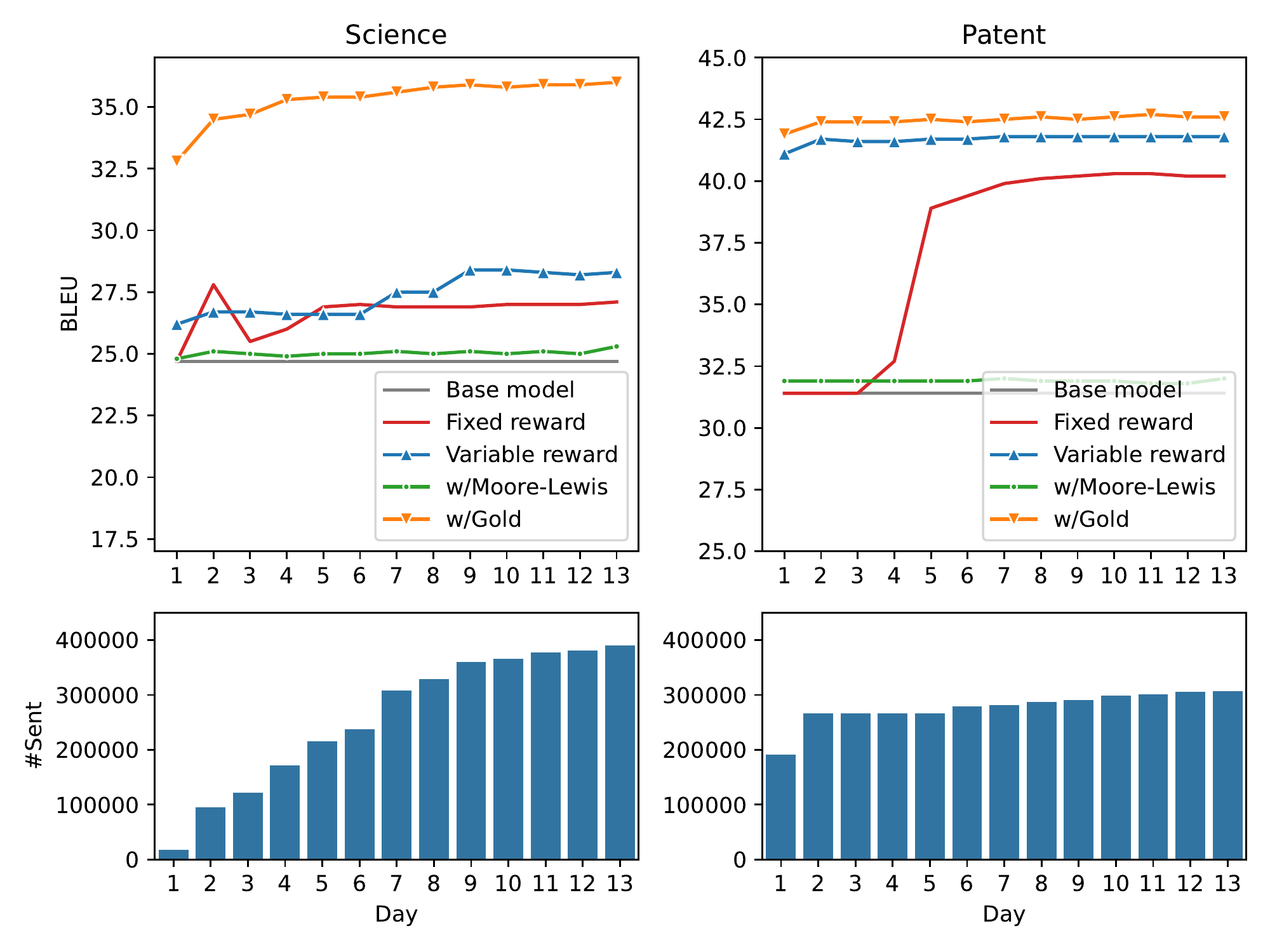}
\caption{
Transitions of BLEU scores (upper-side) and number of collected sentences (lower-side) as we continued data collection for science and patent domains.
Detailed explanations can be found in Section~\ref{sec:large_exp}.
}\label{fig:large_exp_day_sent_bleu_graph_appendix}
\end{figure*}

Fig.~\ref{fig:small_exp_day_sent_graph} shows the numbers of sentences used for fine-tuning in the preliminary experiment (Section~\ref{sec:small_exp}).
Fig.~\ref{fig:large_exp_day_sent_bleu_graph_appendix} shows the transitions of the BLEU scores in the experiment described in Section~\ref{sec:large_exp} and the number of sentences collected in the variable reward setting.

\section{Additional Analysis}
\subsection{Analysis: Reward Function}
\begin{table}[t]
\centering
\scriptsize
\tabcolsep=3pt
\begin{tabular}{lrrr}
\toprule
& {\bf Alignment} & {\bf Domain} & {\bf Both} \\ \midrule
Top 20\% & 34.9 & 34.2 & 34.4 \\
Middle 20\% & 32.9 & 33.3 & 33.1 \\
Bottom 20\% & 30.5 & 32.3 & 31.8 \\
\bottomrule
\end{tabular}
\caption{BLEU scores of model fine-tuned with top/middle/bottom 20\% scored sentences on COVID-19 domain test set}\label{tab:large_exp_bleu_top_reward}
\end{table}

We varied the rewards to the workers with the reward function based on the sentence alignment and domain similarity scores.
We pondered whether this reward function could correctly measure the data quality.
To confirm this, we ordered the collected data with respect to the sentence alignment scores (Eq.~\ref{eq:sentence_alignment}), the domain similarity scores (Eq.~\ref{eq:domain_similarity}), or the sum of both scores.
Then we fine-tuned the model with the top/middle/bottom 20\% of the sorted data.

Table~\ref{tab:large_exp_bleu_top_reward} shows the BLEU scores of the fine-tuned models on the COVID-19 domain.
There is a clear trend that the model fine-tuned with high-scored data achieved higher accuracy, and there is a large gap between the top and the bottom for all the score functions.
From this result, we conclude that our reward function correctly measured the quality of the data, and we paid more for high-quality works and less for low-quality works.

\subsection{Comparison: Gold In-domain Parallel Corpus}
\label{sec:gold}

\begin{table}[t]
\centering
\scriptsize
 \tabcolsep=3pt
\begin{tabular}{lrrrrr}
\toprule
 & \multicolumn{5}{c}{\bf Test BLEU}\\ \cmidrule(lr){2-6}
{\bf Domain} & {\bf Base} & \multicolumn{2}{c}{\bf w/Crawled} & \multicolumn{2}{c}{\bf w/Gold} \\ \midrule
Science & 24.7 & 28.3 \hspace{-0.5em} & (+3.6) & 36.0 \hspace{-0.5em} & (+11.3)\\
Patent & 31.4 & 41.8 \hspace{-0.5em} & (+10.4) & 42.6 \hspace{-0.5em} & (+11.2)\\
Legal & 22.8 & 42.5 \hspace{-0.5em} & (+19.7)  & 59.6 \hspace{-0.5em} & (+36.8)\\
\bottomrule
\end{tabular}
\caption{BLEU score comparison with Gold data (w/Gold)}\label{tab:large_exp_bleu_gold}
\end{table}

In this section, we compare our collected data with the existing domain-specific parallel corpus.
Among the five domains from which we collected sentences, there is a domain-specific parallel corpus for the science, patent, and legal domains.
Note that the availability of domain-specific data is quite limited since creating such parallel data requires professionals, thus incurring heavy costs.
Accordingly, this experiment resembles a comparison between our method and the upper bound.
In the following, we call this domain-specific parallel corpus the gold data.

As gold data, we used ASPEC~\cite{nakazawa16aspec} for the science domain, NTCIR~\cite{goto13ntcir} for the patent domain, and the Japanese-English legal parallel corpus for the legal domain.
For a fair comparison, we randomly selected the same number of sentences as our collected data.

Table~\ref{tab:large_exp_bleu_gold} shows the BLEU scores of the model fine-tuned with the gold data.
Unsurprisingly, the w/Gold models achieved better accuracy than the w/Crawled models.
However, the results of some of the latter were close to those of the former, such as the patent domain.

From Fig.~\ref{fig:large_exp_day_sent_bleu_graph}, we compared the transition of the BLEU scores (see w/Crawled and w/Gold in the legal domain).
From the first to third days, our method’s  performance resembled that of the w/Gold model.
Since room remains for improvements after the fourth day, future work will refine the crowdsourcing protocol.

\section{Links to Data and Software}
\label{sec:appendix_links}
\subsection{Data}
\begin{description}
    \item[JParaCrawl] \url{https://www.kecl.ntt.co.jp/icl/lirg/jparacrawl/}
    \item[TICO-19] \url{https://tico-19.github.io/}
    \item[WMT20 news shared task]\url{http://www.statmt.org/wmt20/translation-task.html}
    \item[ASPEC] \url{http://orchid.kuee.kyoto-u.ac.jp/ASPEC/}
    \item[NTCIR-10] \url{http://research.nii.ac.jp/ntcir/permission/ntcir-10/perm-en-PatentMT.html}
    \item[Legal parallel corpus]\url{http://www.phontron.com/jaen-law/index.html}
\end{description}

\subsection{Software}
\begin{description}
    \item[vecalign] \url{https://github.com/thompsonb/vecalign}
    \item[CLD2] \url{https://github.com/CLD2Owners/cld2}
    \item[KenLM] \url{https://github.com/kpu/kenlm}
    \item[fairseq] \url{https://github.com/pytorch/fairseq}
    \item[SacreBLEU] \url{https://github.com/mjpost/sacreBLEU}
    \item[sentencepiece] \url{https://github.com/google/sentencepiece}
    \end{description}

\end{document}